%% file: acl2018.tex
\documentclass[11pt,a4paper]{article}
\usepackage[hyperref]{acl2018}
\usepackage{times}
\usepackage{latexsym}
\usepackage{amssymb}
\usepackage{amsmath}
\usepackage{booktabs}
\usepackage{multicol}
\usepackage{multirow}
\usepackage{graphicx}
\usepackage{url}
\usepackage{subcaption}

\graphicspath{figures}

\aclfinalcopy 
\def\aclpaperid{293} 

\include{definitions}

\title{\papertitle}

\author{Victor Zhong, Caiming Xiong, Richard Socher \\
  Salesforce Research\\
  Palo Alto, CA \\
  {\tt \{vzhong, cxiong, rsocher\}@salesforce.com}
}
\date{}

\begin{document}
\maketitle


\begin{abstract}
Dialogue state tracking, which estimates user goals and requests given the dialogue context, is an essential part of task-oriented dialogue systems.
In this paper, we propose the \modelname~(\modelnameshort), which learns representations of the user utterance and previous system actions with global-local modules.
Our model uses global modules to share parameters between estimators for different types (called slots) of dialogue states, and uses local modules to learn slot-specific features.
We show that this improves tracking of rare states and achieves \sota~performance on the WoZ and DSTC2 state tracking tasks.
\modelnameshort~obtains \goalacc\% joint goal accuracy and \requestacc\% request accuracy on WoZ, outperforming prior work by \goaldiff\% and \requestdiff\%.
On DSTC2, our model obtains \dstcgoalacc\% joint goal accuracy and \dstcrequestacc\% request accuracy, outperforming prior work by \dstcgoaldiff\% and \dstcrequestdiff\%.
\end{abstract}


\section{Introduction}

\begin{figure}[!t]
\centering
\includegraphics[width=\linewidth]{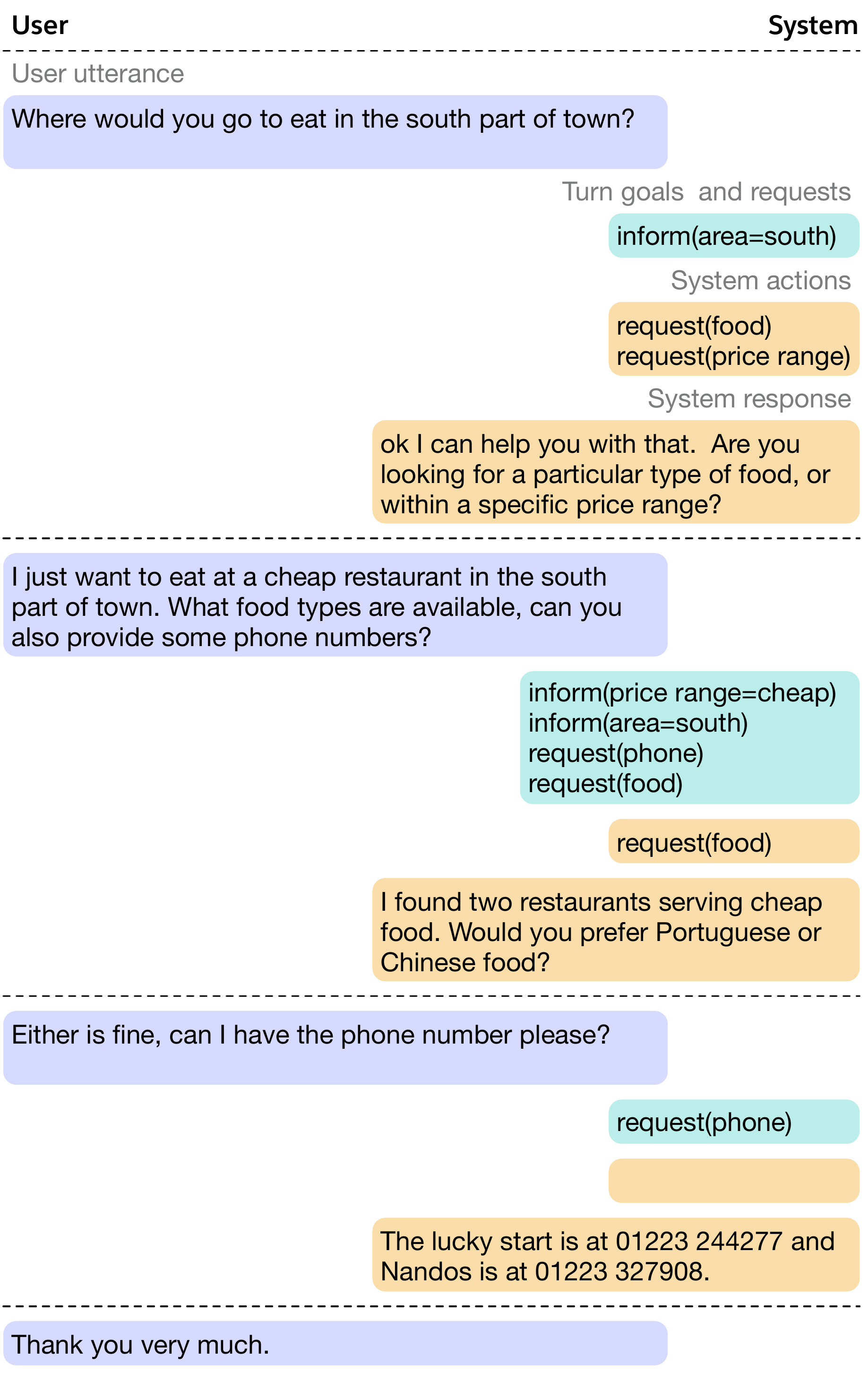}
\caption{
An example dialogue from the WoZ restaurant reservation corpus.
Dashed lines divide turns in the dialogue.
A turn contains an user utterance (purple), followed by corresponding turn-level goals and requests (blue).
The system then executes actions (yellow), and formulates the result into a natural language response (yellow).
}
\label{fig:dialogue}
\end{figure}

Task oriented dialogue systems can reduce operating costs by automating processes such as call center dispatch and online customer support.
Moreover, when combined with automatic speech recognition systems, task-oriented dialogue systems provide the foundation of intelligent assistants such as Amazon Alexa, Apple Siri, and Google Assistant.
In turn, these assistants allow for natural, personalized interactions with users by tailoring natural language system responses to the dialogue context.
Dialogue state tracking (DST) is a crucial part of dialogue systems~\citep{young2013POMDPDialogueReview}.
In DST, a dialogue state tracker estimates the state of the conversation using the current user utterance and the conversation history.
The dialogue system then uses this estimated state to plan the next action and respond to the user.
A state in DST typically consists of a set of \textit{requests} and \textit{joint goals}.
Consider the task of restaurant reservation as an example.
During each turn, the user informs the system of particular \textit{goals} the they would like to achieve (e.g. \texttt{inform(food=french)}), or \textit{request} for more information from the system (e.g. \texttt{request(address)}).
The set of goal and request slot-value pairs (e.g. \texttt{(food, french)}, \texttt{(request, address)}) given during a turn are referred to as the \textit{turn goal} and \textit{turn request}.
The \textit{joint goal} is the set of accumulated turn goals up to the current turn.
Figure~\ref{fig:dialogue} shows an example dialogue with annotated turn states, in which the user reserves a restaurant.

Traditional dialogue state trackers rely on Spoken Language Understanding (SLU) systems~\cite{henderson2012discriminative} in order to understand user utterances.
These trackers accumulate errors from the SLU, which sometimes do not have the necessary dialogue context to interpret the user utterances.
Subsequent DST research forgo the SLU and directly infer the state using the conversation history and the user utterance~\cite{henderson2014word,zilka2015incremental,mrkvsic2015multi}.
These trackers rely on hand-crafted semantic dictionaries and delexicalization --- the anonymization of slots and values using generic tags --- to achieve generalization.
Recent work by~\citet{mrkvsic2016neural} apply representation learning using convolutional neural networks to learn features relevant for each state as opposed to hand-crafting features.


A key problem in DST that is not addressed by existing methods is the extraction of rare slot-value pairs that compose the state during each turn.
Because task oriented dialogues cover large state spaces, many slot-value pairs that compose the state rarely occur in the training data.
Although the chance that the user specifies a particular rare slot-value in a turn is small, the chance that they specify at least one rare slot-value pair is large.
Failure to predict these rare slot-value pairs results in incorrect turn-level goal and request tracking.
Accumulated errors in turn-level goal tracking significantly degrade joint goal-tracking.
For example, in the WoZ state tracking dataset, slot-value pairs have 214.9 training examples on average, while 38.6\% of turns have a joint goal that contains a rare slot-value pair with less than 20 training examples.

In this work, we propose the \modelnamehighlight~(\modelnameshort), a new \sota~ model for dialogue state tracking.
In contrast to previous work that estimate each slot-value pair independently,
\modelnameshort~ uses global modules to share parameters between estimators for each slot and local modules to learn slot-specific feature representations.
We show that by doing so, \modelnameshort~generalizes on rare slot-value pairs with few training examples.
\modelnameshort~achieves \sota~results of \goalacc\% goal accuracy and \requestacc\% request accuracy on the WoZ dialogue state tracking task~\citep{wen2017NetworkBasedEndToEndDialogueSystem}, outperforming prior best by \goaldiff\% and \requestdiff\%.
On DSTC2~\citep{dstc2}, we achieve \dstcgoalacc\% goal accuracy and \dstcrequestacc\% request accuracy, outperforming prior best by \dstcgoaldiff\% and \dstcrequestdiff\%.
We release an implementation of our model along with a Docker image of our experiment setup for reproducibility~\footnote{https://github.com/salesforce/glad}.


\section{\modelname}

\begin{figure*}[!t]
\centering
\includegraphics[width=\linewidth]{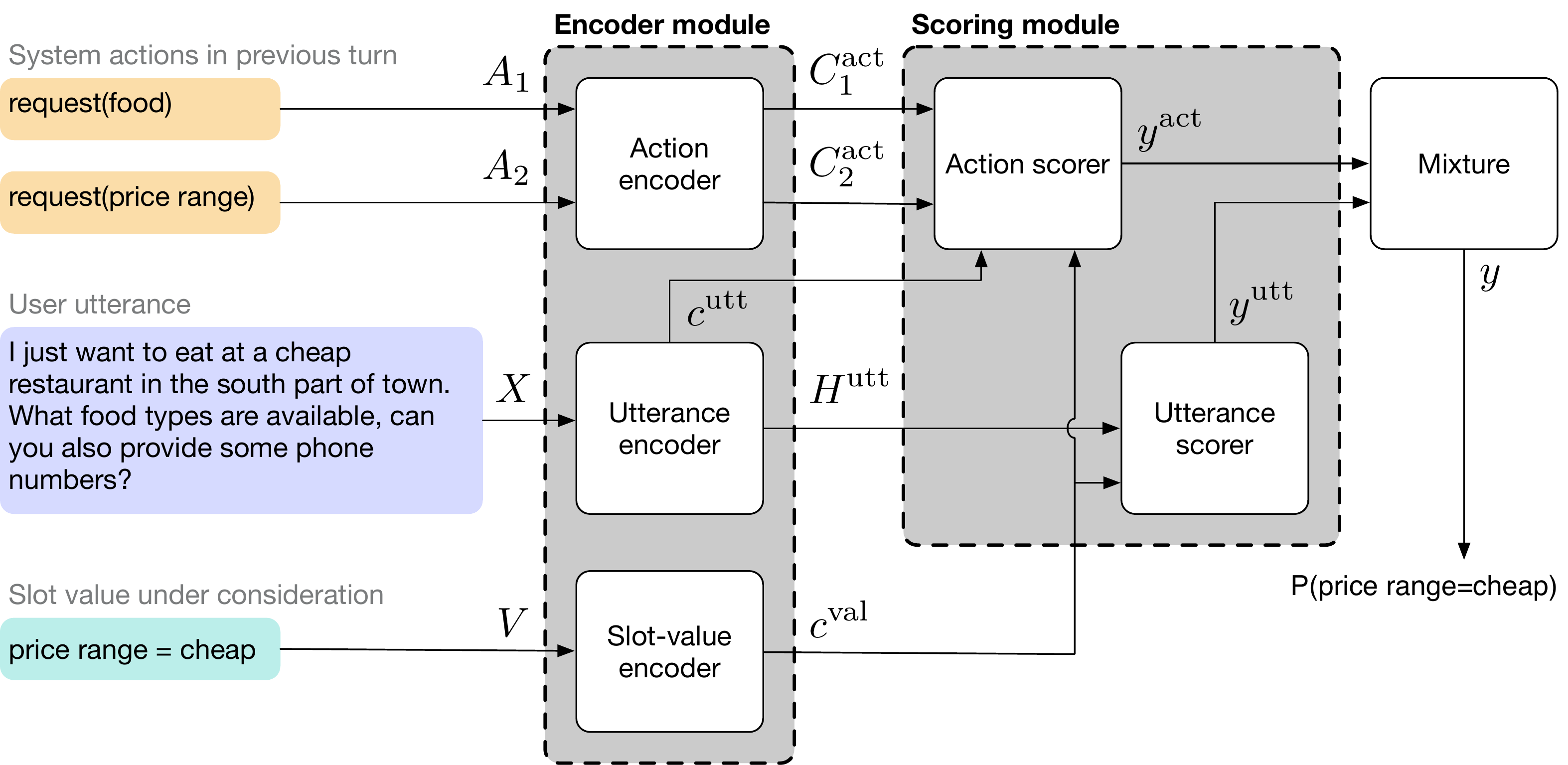}
\caption{
The \modelname.
}
\label{fig:model}
\end{figure*}

One formulation of state tracking is to predict the turn state given an user utterance and previous system actions~\citep{williams2007partially}.
Like previous methods~\citep{henderson2014word,wen2017NetworkBasedEndToEndDialogueSystem,mrkvsic2016neural}, \modelnameshort~decomposes the multi-label state prediction problem into a collection of binary prediction problems by using a distinct estimator for each slot-value pair that make up the state.
Hence, we describe \modelnameshort~with respect to a slot-value pair that is being predicted by the model.

Shown in Figure~\ref{fig:model}, \modelnameshort~is comprised of an encoder module and a scoring module.
The encoder module consists of separate global-locally self-attentive encoders for the user utterance, the previous system actions, and the slot-value pair under consideration.
The scoring module consists of two scorers.
One scorer considers the contribution from the utterance while the other considers the contribution from previous system actions.

\subsection{Global-Locally Self-Attentive Encoder}

\begin{figure}[t]
\centering
\includegraphics[width=\linewidth]{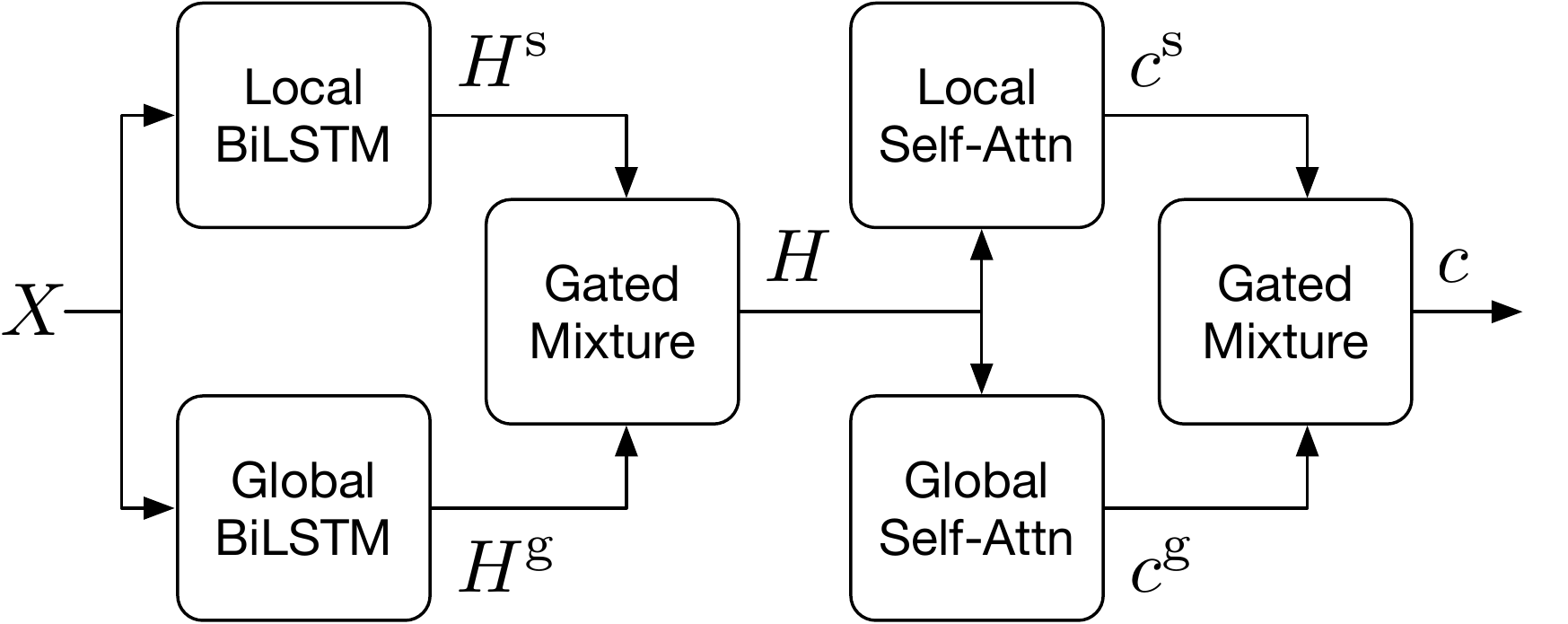}
\caption{
Global-locally self-attentive encoder.
}
\label{fig:share}
\end{figure}

We begin by describing the global-locally self-attentive encoder, which makes up the encoder module.
DST datasets tend to be small relative to their state space in that many slot-value pairs rarely occur in the dataset.
Because each state is comprised of a set of slot-value pairs, many of them rare, poor inference of rare slot-value pairs subsequently results in poor turn-level tracking.
This problem is amplified in joint tracking, due to the accumulation of turn-level errors.
In developing this encoder, we seek to better model rare slot-value pairs by sharing parameters between each slot through global modules and learning slot-specific features through local modules.

The global-locally self-attentive encoder consists of a bidirectional LSTM~\citep{Hochreiter1997Long}, which captures temporal relationships within the sequence, followed by a self-attention layer to compute the summary of the sequence.
Figure~\ref{fig:share} illustrates the global-locally self-attentive encoder.

Consider the process of encoding a sequence with respect to a particular slot $s$.
Let $\ngeneral$ denote the number of words in the sequence, $\demb$ the dimension of the embeddings, and $X \in \real{\ngeneral \times \demb}$ the word embeddings corresponding to words in the sequence.
We produce a global encoding $H\vglobal$ of $X$ using a global bidirectional LSTM.

\begin{eqnarray}
H\vglobal &=& \bilstm\vglobal \left( X \right) \in \real{\ngeneral \times \drnn}
\end{eqnarray}

where $\drnn$ is the dimension of the LSTM state.
We similarly produce a local encoding $H\vlocal$ of $X$, taking into account the slot $s$, using a local bidirectional LSTM.

\begin{eqnarray}
H\vlocal &=& \bilstm\vlocal \left( X \right) \in \real{\ngeneral \times \drnn}
\end{eqnarray}

The outputs of the two LSTMs are combined through a mixture function to yield a global-local encoding $H$ of $X$. 

\begin{eqnarray}
H = \beta\vlocal H\vlocal + \left( 1 - \beta\vlocal \right) H\vglobal \in \real{\ngeneral \times \drnn}
\end{eqnarray}

Here, the scalar $\beta\vlocal$ is a learned parameter between 0 and 1 that is specific to the slot $s$.
Next, we compute a global-local self-attention context $c$ over $H$.
Self-attention, or intra-attention, is an effective method to compute summary context over variable-length sequences for natural language processing tasks~\citep{cheng2016long,Vaswani2017attention,he2017deep,lee2017end}.
In our case, we use a global self-attention module to compute an attention context useful for general-purpose state tracking, as well as a local self-attention module to compute a slot-specific attention context.

For each $i$th element $H_i$, we compute a scalar global self-attention score $a\vglobal_i$ which is subsequently normalized across all elements using a softmax function.

\begin{eqnarray}
\label{eq:selfattnscore}
a\vglobal_i &=& W\vglobal H_i + b\vglobal \in \real{} \\
p\vglobal &=& \softmax \left( a\vglobal \right) \in \real{\ngeneral}
\end{eqnarray}

The global self-attention context $c\vglobal$ is then the sum of each element $H_i$, weighted by the corresponding normalized global self-attention score $p\vglobal_i$.

\begin{eqnarray}
\label{eq:selfattncontext}
c\vglobal &=& \sum_i p\vglobal_i H_i \in \real{\drnn}
\end{eqnarray}

We similarly compute the local self-attention context $c\vlocal$.

\begin{eqnarray}
a\vlocal_i &=& W\vlocal H_i + b\vlocal \in \real{} \\
p\vlocal &=& \softmax \left( a\vlocal \right) \in \real{\ngeneral} \\
c\vlocal &=& \sum_i p\vlocal_i H_i \in \real{\drnn}
\end{eqnarray}

The global-local self-attention context $c$ is the mixture

\begin{eqnarray}
c = \beta\vlocal c\vlocal + \left( 1 - \beta\vlocal \right) c\vglobal \in \real{\drnn}
\end{eqnarray}

For ease of exposition, we define the multi-value encode function $\encode \left( X \right)$.

\begin{eqnarray}
\encode : X \rightarrow H, c
\end{eqnarray}

This function maps the sequence $X$ to the encoding $H$ and the self-attention context $c$.

\subsection{Encoding module}

Having defined the global-locally self-attentive encoder, we now build representations for the user utterance, the previous system actions, and the slot-value pair under consideration.
Let $U$ denote word embeddings of the user utterance, $A_j$ denote those of the $j$th previous system action (e.g. \texttt{request ( price range )}, and $V$ denote those of the slot-value pair under consideration (e.g. \texttt{food = french}).
We have

\begin{eqnarray}
H\utt, c\utt &=& \encode \left( U \right) \\
H\act_j, C\act_j &=& \encode \left( A_j \right) \\
H\val, c\val &=& \encode \left( V \right) 
\end{eqnarray}

\subsection{Scoring module}

Intuitively, we can determine whether the user has expressed the slot-value pair under consideration by examining two input sources.
The first source is the user utterance, in which the user directly states the goals and requests.
An example of this is the user saying ``how about a French restaurant in the centre of town?'', after the system asked ``how may I help you?''
To handle these cases, we determine whether the utterance specifies the slot-value pair.
Namely, we attend over the user utterance $H\utt$, taking into account the slot-value pair being considered $c\val$, and use the resulting attention context $q\utt$ to score the slot-value pair.

\begin{eqnarray}
a\utt_i &=& \left( H\utt_i \right)^\intercal c\val \in \real{}\\
p\utt &=& \softmax \left( a\utt \right) \in \real{\nutterance}\\
q\utt &=& \sum_i p\utt_i H\utt_i \in \real{\drnn} \\
\label{eq:uttscore}
y\utt &=& W q\utt + b \in \real{}
\end{eqnarray}

where $\nutterance$ is the number of words in the user utterance.
The score $y\utt$ indicates the degree to which the value was expressed by the user utterance.

The second source is the previous system actions.
This source is informative when the user utterance does not present enough information and instead refers to previous system actions.
An example of this is the user saying ``yes'', after the system asked ``would you like a restaurant in the centre of town?''
To handle these cases, we examine previous actions after considering the user utterance.
First, we attend over the previous action representations $C\act = [C\act_1 \cdots C\act_\naction]$, taking into account the current user utterance $c\utt$.
Here, $\naction$ is the number of previous system actions.
Then, we use the similarity between the attention context $q\act$ and the slot-value pair $c\val$ to score the slot-value pair.

\begin{eqnarray}
a\act_j &=& \left( C\act_j \right) ^ \intercal c\utt \in \real{} \\
p\act &=& \softmax \left( a\act \right) \in \real{\naction + 1} \\
q\act &=& \sum_j p\act_j C\act_j \in \real{\drnn} \\
\label{eq:actscore}
y\act &=& \left( q\act \right) ^ \intercal c\val \in \real{}
\end{eqnarray}

In addition to real actions, we introduce a sentinel action to each turn which allows the attention mechanism to ignore previous system actions.
The score $y\act$ indicates the degree to which the value was expressed by the previous actions.

The final score $y$ is then a weighted sum between the two scores $y\utt$ and $y\act$, normalized by the sigmoid function $\sigma$.

\begin{eqnarray}
\label{eq:score}
y &=& \sigmoid \left( y\utt + w y\act \right) \in \real{}
\end{eqnarray}

Here, the weight $w$ is a learned parameter.

\section{Experiments}

\begin{table*}[t]
\centering
\begin{tabular}{lrrrr}
\toprule
\multirow{2}{*}{Model}                             & \multicolumn{2}{c}{DSTC2} & \multicolumn{2}{c}{WoZ} \\
                                                   & Joint goal   & Turn request  & Joint goal  & Turn request \\
\midrule
Delexicalisation-Based Model                       & 69.1\%       & 95.7\%         & 70.8\%      & 87.1\%        \\
Delex. Model + Semantic Dictionary                 & 72.9\%       & 95.7\%         & 83.7\%      & 87.6\%        \\
Neural Belief Tracker (NBT) - DNN                        & 72.6\%       & 96.4\%         & 84.4\%      & 91.2\%        \\
Neural Belief Tracker (NBT) - CNN                        & 73.4\%       & 96.5\%         & 84.2\%      & 91.6\%        \\
\modelnameshort                                         & \textbf{\dstcgoalacc $\pm$ \dstcgoalstd\%}       & \textbf{\dstcrequestacc $\pm$ \dstcrequeststd\%}         & \textbf{\goalacc $\pm$ \goalstd\%}      & \textbf{\requestacc $\pm$ \requeststd\%}        \\
\bottomrule
\end{tabular}
\caption{
Test accuracies on the DSTC2 and WoZ restaurant reservation datasets.
The other models are: delexicalisation DSTC2~\citep{henderson2014word}, delexicalisation WoZ~\citep{wen2017NetworkBasedEndToEndDialogueSystem}, and NBT~\citep{mrkvsic2016neural}.
We run 10 models using random seeds with early stopping on the development set, and report the mean and standard deviation test accuracies for each dataset.
}
\label{tb:result}
\vspace{-0.2cm}
\end{table*}

\subsection{Dataset}
The Dialogue Systems Technology Challenges (DSTC) provides a common framework for developing and evaluating dialogue systems and dialogue state trackers~\cite{dstc1,dstc2}.
Under this framework, dialogue semantics such as states and actions are based on a task ontology such as restaurant reservation.
During each turn, the user informs the system of particular \textit{goals} (e.g. \texttt{inform(food=french)}), or \textit{requests} for more information from the system (e.g. \texttt{request(address)}).
For instance, \texttt{food} and \texttt{area} are examples of slots in the DSTC2 task, and \texttt{french} and \texttt{chinese} are example values within the \texttt{food} slot.
We train and evaluate our model using DSTC2 as well as the Wizard of Oz (WoZ) restaurant reservation task~\cite{wen2017NetworkBasedEndToEndDialogueSystem}, which also adheres to the DSTC framework and has the same ontology as DSTC2.

For DSTC2, it is standard to evaluate using the N-best list of the automatic speech recognition system (ASR) that is included with the dataset.
Because of this, each turn in the DSTC2 dataset contains several noisy ASR outputs instead of a noise-free user utterance.
The WoZ task does not provide ASR outputs, and we instead train and evaluate using the user utterance.

\subsection{Metrics}
We evaluate our model using turn-level request tracking accuracy as well as joint goal tracking accuracy.
Our definition of \modelnameshort~in the previous sections describes how to obtain turn goals and requests.
To compute the joint goal, we simply accumulate turn goals.
In the event that the current turn goal specifies a slot that has been specified before, the new specification takes precedence.
For example, suppose the user specifies a \texttt{food=french} restaurant during the current turn. 
If the joint goal has no existing \texttt{food} specifications, then we simply add \texttt{food=french} to the joint goal.
Alternatively, if \texttt{food=thai} had been specified in a previous turn, we simply replace it with \texttt{food=french}.

\subsection{Implementation Details}
We use fixed, pretrained GloVe embeddings~\cite{pennington2014glove} as well as character n-gram embeddings~\cite{Hashimoto2017joint}.
Each model is trained using ADAM~\cite{kingma2014adam}.
For regularization, we apply dropout with 0.2 drop rate~\cite{srivastava2014dropout} to embeddings and the output of each local and global module.
We use the development split for hyperparameter tuning and apply early stopping using the joint goal accuracy.


For the DSTC2 task, we train using transcripts of user utterances and evaluate using the noisy ASR transcriptions.
During evaluation, we take the sum of the scores resulting from each ASR output as the output score of a particular slot-value.
We then normalize this sum using a sigmoid function as shown in Equation~\eqref{eq:score}.
We also apply word dropout, in which the embeddings of a word is randomly set to zero with a probability of 0.3.
This accounts for the poor quality of ASR outputs in DSTC2, which frequently miss words in the user utterance.
We did not find word dropout to be helpful for the WoZ task, which does not contain noisy ASR outputs.

\subsection{Comparison to Existing Methods}

Table~\ref{tb:result} shows the performance of \modelnameshort~compared to previous \sota~models.
The delexicalisation models, which replace slots and values in the utterance with generic tags, are from~\citet{henderson2014word} for DSTC2 and~\citet{wen2017NetworkBasedEndToEndDialogueSystem} for WoZ.
Semantic dictionaries map slot-value pairs to hand-engineered synonyms and phrases.
The NBT~\citep{mrkvsic2016neural} applies CNN over word embeddings learned over a paraphrase database~\citep{wieting2015paraphrase} instead of delexicalised n-gram features.

On the WoZ dataset, we find that \modelnameshort~significantly improves upon previous \sota~performance by \goaldiff\% on joint goal tracking accuracy and \requestdiff\% on turn-level request tracking accuracy.
On the DSTC dataset, which evaluates using noisy ASR outputs instead of user utterances, \modelnameshort~improves upon previous state of the art performance by \dstcgoaldiff\% on joint goal tracking accuracy and \dstcrequestdiff\% on turn-level request tracking accuracy.

\begin{table}[t]
\centering
\begin{tabular}{lrrr}
\toprule
Model                    & Tn goal & Jnt goal & Tn request \\
\midrule
\modelnameshort          & 93.7\%  & 88.8\%  & 97.3\%  \\
- global         & 88.8\%  & 73.4\%  & 97.3\%  \\
- local          & 93.1\%  & 86.6\%  & 95.1\%  \\
- self-attn         & 91.6\%  & 84.4\%  & 97.1\%  \\
- LSTM                   & 88.7\%  & 71.5\%  & 93.2\%  \\
\bottomrule
\end{tabular}
\caption{
Ablation study showing turn goal, joint goal, and turn request accuracies on the dev. split of the WoZ dataset.
For ``- self-attn'', we use mean-pooling instead of self-attention.
For ``- LSTM'', we compute self-attention over word embeddings.
}
\label{tb:ablation}
\vspace{-0.2cm}
\end{table}

\subsection{Ablation study}

We perform ablation experiments on the development set to analyze the effectiveness of different components of \modelnameshort.
The results of these experiments are shown in Table~\ref{tb:ablation}.
We also show turn goal accuracy in addition to joint goal accuracy and turn request accuracy for reference.

\textbf{Temporal order is important for state tracking.}
We experiment with an embedding-matching variant of \modelnameshort~with self-attention but without LSTMs.
The weaker performance by this model suggests that representations that capture temporal dependencies is helpful for understanding phrases for state tracking.

\begin{figure*}[t]
\centering
\includegraphics[width=0.9\linewidth]{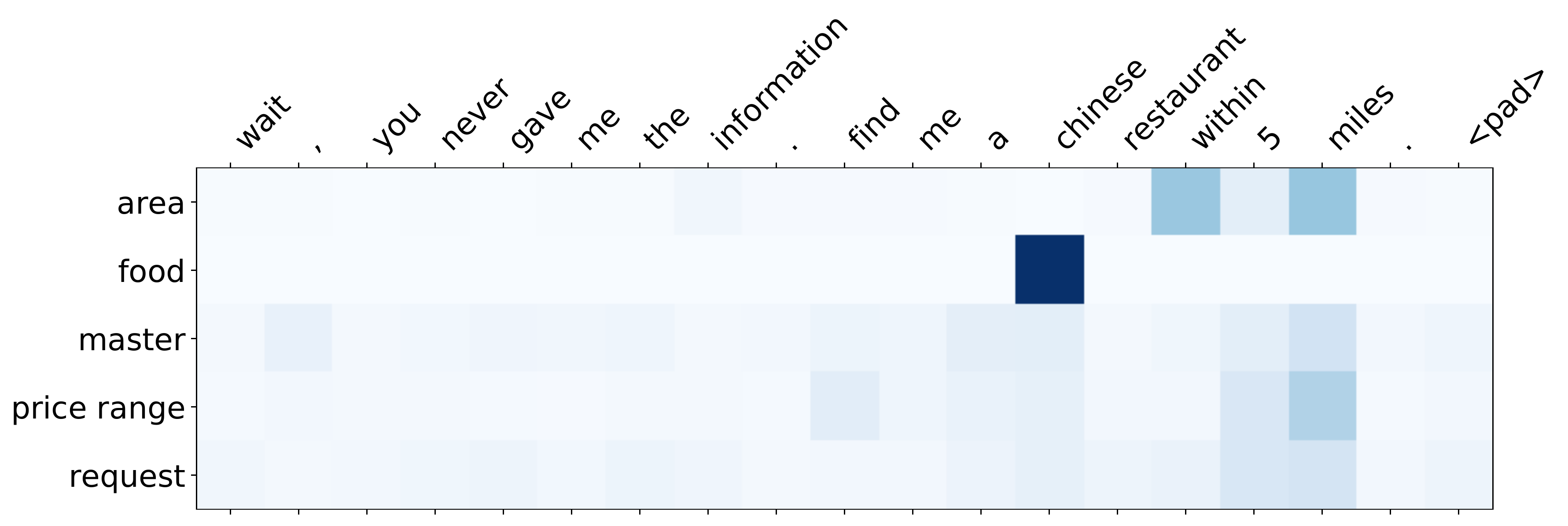}
\caption{
Global and local self-attention scores on user utterances.
Each row corresponds to the self-attention score for a particular slot.
Slot-specific local self-attention modules emphasize relevant key words and phrases to that slot, whereas the global module attends to all relevant words.
}
\label{fig:selfattn}
\vspace{-0.2cm}
\end{figure*}

\textbf{Self-attention allows slot-specific, robust feature learning.}
We observe a consistent drop in performance for models that use mean-pooling over the temporal dimension as opposed to self-attention (Equations~\eqref{eq:selfattnscore} to~\eqref{eq:selfattncontext}).
This stems from the flexibility in the attention context computation afforded by the self-attention mechanism, which allows the model to focus on select words relevant to the slot-value pair under consideration.
Figure~\ref{fig:selfattn} illustrates an example in which local self-attention modules focus on relevant parts of the utterance.
The model attends to relevant phrases that n-gram and embedding matching techniques do not capture (e.g. ``within 5 miles'' for the ``area'' slot).

\begin{figure}[t]
\centering
\includegraphics[width=\linewidth]{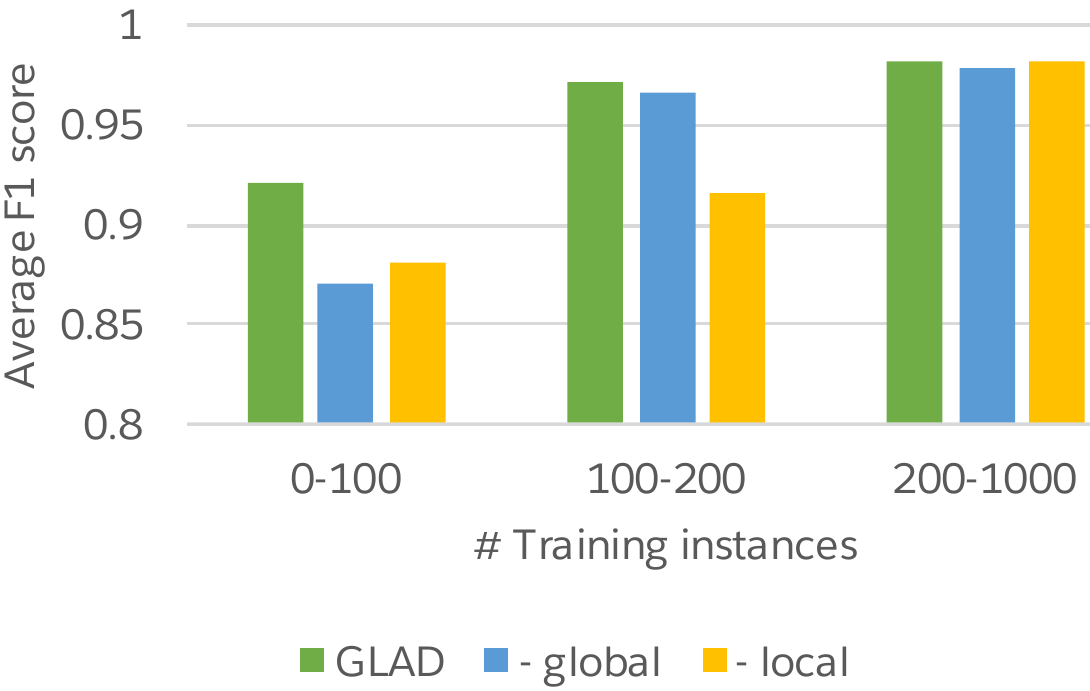}
\caption{
F1 performance for each slot-value pair in the development split of the WoZ task, grouped by the number of training instances.
}
\label{fig:rarity}
\vspace{-0.2cm}
\end{figure}

\textbf{Global-local sharing improves goal tracking.}
We study the two extremes of sharing between the global module and the local module.
The first uses only the local module and results in degradation in goal tracking but does not affect request tracking (e.g. $\beta^s = 1$).
This is because the former is a joint prediction over several slot-values with few training examples, whereas the latter predicts a single slot that has the most training examples.

The second uses only the global module and underperforms in both goal tracking and request tracking (e.g. $\beta^s = 0$).
This model is less expressive, as it lacks slot-specific specializations except for the final scoring modules.

Figure~\ref{fig:rarity} shows the performance of~\modelnameshort~and the two sharing variants across different numbers of occurrences in the training data.
\modelnameshort~consistently outperforms both variants for rare slot-value pairs.
For slot-value pairs with an abundance of training data, there is no significant performance difference between models as there is sufficient data to generalize.

\subsection{Qualitative analysis}

Table~\ref{tb:preds} shows example predictions by \modelnameshort.
In the first example, the user explicitly outlines requests and goals in a single utterance.
In the second example, the model previously prompted the user for confirmation of two requests (e.g. for the restaurant's address and phone number), and the user simply answers in the affirmative.
In this case, the model obtains the correct result by leveraging the system actions in the previous turn.
The last example demonstrates an error made by the model.
Here, the user does not answer the system's previous request for the choice of food and instead asks for what food is available.
The model misinterprets the lack of response as the user not caring about the choice of food.

\section{Related Work}

\input{preds}

\textbf{Dialogue State Tracking.}
Traditional dialogue state trackers rely on a separate SLU component that serves as the initial stage in the dialogue agent pipeline.
The downstream tracker then combines the semantics extracted by the SLU with previous dialogue context in order to estimate the current dialogue state~\cite{thomson2010bayesian,wang2013simple,williams2014web,perez2016dialog}.
Recent results in dialogue state tracking show that it is beneficial to jointly learn speech understanding and dialogue tracking~\cite{henderson2014word,zilka2015incremental,wen2017NetworkBasedEndToEndDialogueSystem}.
These approaches directly take as input the N-best list produced by the ASR system.
By avoiding the accumulation of errors from the initial SLU component, these joint approaches typically achieved stronger performance on tasks such as DSTC2.
One drawback to these approaches is that they rely on hand-crafted features and complex domain-specific lexicon (in addition to the ontology), and consequently are difficult to extend and scale to new domains.
The recent Neural Belief Tracker (NBT) by~\citet{mrkvsic2016neural} avoids reliance on hand-crafted features and lexicon by using representation learning.
The NBT employs convolutional filters over word embeddings in lieu of previously-used hand-engineered features.
Moreover, to outperform previous methods, the NBT uses pretrained embeddings tailored to retain semantic relationships by injecting semantic similarity constraints from the Paraphrase Database~\cite{wieting2015paraphrase,ganitkevitch2013ppdb}.
On the one hand, these specialized embeddings are more difficult to obtain than word embeddings from language modeling.
On the other hand, these embeddings are not specific to any dialogue domain and generalize to new domains.

\textbf{Neural attention models in NLP.}
Attention mechanisms have led to improvements on a variety of natural language processing tasks.
\citet{Bahdanau2014NeuralMT} propose attentional sequence to sequence models for neural machine translation.
\citet{luong2015effective} analyze various attention techniques and highlight the effectiveness of the simple, parameterless dot product attention.
Similar models have also proven successful in tasks such as summarization~\cite{see2017get,paulus2017deep}.
Self-attention, or intra-attention, has led improvements in language modeling, sentiment analysis, natural language inference~\cite{cheng2016long}, semantic role labeling~\cite{he2017deep}, and coreference resolution~\citep{lee2017end}.
Deep self-attention has also achieved \sota~results in machine translation~\cite{Vaswani2017attention}.
Coattention, or bidirectional attention that codependently encode two sequences, have led to significant gains in question answering~\cite{Xiong2016dynamic,seo2016bidirectional,xiong2018dcn} as well as visual question answering~\cite{lu2016hierarchical}.

\textbf{Parameter sharing between related tasks.}
Sharing parameters between related tasks to improve joint performance is prominent in multi-task learning~\cite{caruana1998multitask,thrun1996learning}.
Early works in multi-tasking use Gaussian processes whose covariance matrix is induced from shared kernels~\cite{lawrence2004learning,yu2005learning,seeger2005semiparametric,bonilla2008multi}.
\citet{Hashimoto2017joint} propose a progressively trained joint model for NLP tasks.
When a new task is introduced, a new section is added to the network whose inputs are intermediate representations from sections for previous tasks.
In this sense, tasks share parameters in a hierarchical manner.
\citet{johnson2016google} propose a single model that jointly learns to translate between multiple language pairs, including one-to-many, many-to-one, and many-to-many translation.
\citet{kaiser2017one} propose a model that jointly learns multiple tasks across modalities.
Each modality-specific feature extractor extracts a representation that is fed into a shared encoder.






\section{Conclusions}
We introduced the \modelname~(\modelnameshort), a new \sota~ model for dialogue state tracking.
At the core of \modelnameshort~is the global-locally self-attention encoder, whose global modules allow parameter sharing between slots and local modules allow slot-specific feature learning.
This allows \modelnameshort~to generalize on rare slot-value pairs with few training data.
\modelnameshort~achieves \sota~results of \goalacc\% goal accuracy and \requestacc\% request accuracy on the WoZ dialogue state tracking task, as well as \dstcgoalacc\% goal accuracy and \dstcrequestacc\% request accuracy on DSTC2.

\section*{Acknowledgement}
We thank Nikola Mrk\v{s}i\'{c} for helpful discussion and for providing a preprocessed version of the DSTC2 dataset.

\bibliography{acl2018}
\bibliographystyle{acl}

%

\end{document}

%% file: definitions.tex
\newcommand{\modelname}{Global-Locally Self-Attentive Dialogue State Tracker}
\newcommand{\modelnamehighlight}{\textbf{G}lobal-\textbf{L}ocally Self-\textbf{A}ttentive \textbf{D}ialogue State Tracker}
\newcommand{\modelnameshort}{GLAD}
\newcommand{\papertitle}{Global-Locally Self-Attentive Dialogue State Tracker}

\newcommand{\sota}{state-of-the-art}

\newcommand{\demb}{d_{\rm{emb}}}

\newcommand{\drnn}{d_{\rm{rnn}}}

\newcommand{\ngeneral}{n}
\newcommand{\nutterance}{m}
\newcommand{\naction}{l}

\newcommand{\utt}{^{\rm{utt}}}
\newcommand{\act}{^{\rm{act}}}
\newcommand{\val}{^{\rm{val}}}

\newcommand{\vglobal}{^{\rm{g}}}
\newcommand{\vlocal}{^{\rm{s}}}

\newcommand{\goaldiff}{3.7}  
\newcommand{\requestdiff}{5.5}  

\newcommand{\goalacc}{88.1}
\newcommand{\requestacc}{97.1}
\newcommand{\goalstd}{0.4}
\newcommand{\requeststd}{0.2}

\newcommand{\dstcgoaldiff}{1.1}  
\newcommand{\dstcrequestdiff}{1.0}  
\newcommand{\dstcgoalacc}{74.5}
\newcommand{\dstcrequestacc}{97.5}
\newcommand{\dstcgoalstd}{0.2}
\newcommand{\dstcrequeststd}{0.1}

\newcommand{\real}[1]{\mathbb{R}^{#1}}
\newcommand{\bilstm}{{\rm{biLSTM}}}
\newcommand{\sigmoid}{{\sigma}}
\newcommand{\softmax}{{\rm{softmax}}}

\newcommand{\encode}{{\rm{encode}}}

%% file: preds.tex
\begin{table*}[t]
\centering
\begin{tabular}{@{}p{6.5cm}p{4.3cm}l@{}}
\toprule

System actions in previous turn
&
User utterance
&
Predicted turn belief state
\\ \midrule


N/A
&
I would like Polynesian food in the South part of town. Please send me phone number and address.
&
\begin{tabular}[t]{@{}l@{}}
request(phone)\\ request(address)\\ inform(food=polynesian)\\ inform(area=south)
\end{tabular}
\\ \midrule


\begin{tabular}[t]{@{}p{6.5cm}@{}}
request(address)\\ request(phone)
\\\\
There is a moderately priced italian place called Pizza hut at cherry hilton. would you like the address and phone number?
\end{tabular}
&
Yes please.
&
\begin{tabular}[t]{@{}l@{}}
request(phone)\\ request(address)
\end{tabular}
\\ \midrule


\begin{tabular}[t]{@{}p{6.5cm}@{}}
request(food)\\ request(price range)
\\
\\
ok I can help you with that. Are you looking for a particular type of food, or within a specific price range?
\end{tabular}
&
I just want to eat at a cheap restaurant in the south part of town. What food types are available, can you also provide some phone numbers?
&
\begin{tabular}[t]{@{}l@{}}
  request(phone)\\ inform(price range=cheap)\\ inform(area=south)\\ {\color{red} -inform(food=dontcare)}\\ {\color{blue} +request(food)}
\end{tabular}

\\ \bottomrule
\end{tabular}%
\caption{
Example predictions by \modelname~on the development split of the WoZ restaurant reservation dataset.
Model predicted slot-value pairs that are not in the ground truth (e.g. {\color{blue} false positives}) are prefaced with a ``{\color{blue} +}'' symbol.
Ground truth slot-value pairs that are not predicted by the model (e.g. {\color{red} false negatives}) are prefaced with a ``{\color{red} -}'' symbol.
}
\label{tb:preds}
\vspace{-0.3cm}
\end{table*}